\documentclass[runningheads]{llncs}
\usepackage{graphicx}
\usepackage{booktabs}

\begin{document}
\title{Exploring Deep Learning Models for EEG Neural Decoding}
\titlerunning{Exploring Deep Learning Models for EEG Neural Decoding}
%
\author{Laurits Dixen \and Stefan Heinrich \and Paolo Burelli}
\authorrunning{Dixen et al.}
\institute{IT University of Copenhagen, Denmark}

\maketitle 
\begin{abstract}
Neural decoding is an important method in cognitive neuroscience that aims to decode brain representations from recorded neural activity using a multivariate machine learning model. The THINGS initiative \cite{grootswagersHumanEEGRecordings2022} provides a large EEG dataset of 46 subjects watching rapidly shown images. Here, we test the feasibility of using this method for decoding high-level object features using recent deep learning models. We create a derivative dataset from this of living vs non-living entities test 15 different deep learning models with 5 different architectures and compare to a SOTA linear model. We show that the linear model is not able to solve the decoding task, while almost all the deep learning models are successful, suggesting that in some cases non-linear models are needed to decode neural representations. We also run a comparative study of the models' performance on individual object categories, and suggest how artificial neural networks can be used to study brain activity.

\end{abstract}

 
\section{Introduction}

The human brain is still the most advanced information-processing entity we know of, with incredible efficiency, efficacy and adaptability. However, the details of how information is represented and processed in the brain at the algorithmic level are elusive to us. Furthering this understanding holds great potential for understanding intelligence at a deeper level. Neural decoding is a somewhat recent technique for studying neural representations. It consists of applying a machine learning (ML) model to recordings of brain activity, to decode representations in the brain \cite{kriegeskorteInterpretingEncodingDecoding2019}. Here decoding is finding dissociations in the neural activity associated with different stimuli or conditions. When dissociation is successful, that neural activity is taken to contain relevant information about the conditional difference, thereby giving scientists a clue on what information is present where and when in coganitive processing. Neural decoding, using multivariate statistical or ML approaches, has gained popularity among cognitive neuroscientists as a tool to analyse their experimental data, replacing, to some degree, the traditional approach of inferential statistics on univariate measures of individual voxels or electrodes using generalised linear models \cite{frisbyDecodingSemanticRepresentations2023}. Recently, neural decoding has given great insight into among others vision science \cite{robinsonVisualRepresentationsInsights2023}, face processing \cite{kanwisherCNNsRevealComputational2023}. The imaging technique focused on in this paper is electroencephalography (EEG). EEG offers a non-invasive way to measure neural activity with a very high sampling rate at the cost of a lower spatial resolution when compared to other popular imaging techniques. Whereas high accuracy in traditional neural decoding is not essential - if decoding happens reliably it is taken as such, other fields using similar techniques have great interest in creating models able to efficiently learn associations between brain patterns in EEG and behaviour, intention or cognitive state \cite{robinsonVisualRepresentationsInsights2023}. Among these fields is the study of brain-computer interfaces (BCI), where the aim is to allow interactions with a computer using only neural recordings. Medical fields are also starting to rely on using computational analysis of brain activity to diagnose neurological diseases such as depression\cite{thoduparambilEEGbasedDeepLearning2020} and dementia \cite{kimDeepLearningbasedEEG2023}. Other fields that also benefit from high-performing models on neural data are affective computing \cite{koelstraDEAPDatabaseEmotion2012}, and stimulus reconstruction \cite{benchetritBrainDecodingRealtime2024}. Therefore, developing models that can effectively learn the patterns in EEG signals is highly important. Even for traditional cognitive neural decoding, simpler models might not be able to successfully find some associations that only show in highly nonlinear relationships \cite{girnComplexSystemsPerspective2023a}.

Deep learning (DL) models have, especially in the last 10 years, shown their power in modelling nonlinear relationships. Even so much that they have gotten to human or superhuman levels in some cognitively relevant tasks. This has caused interest from cognitive neuroscientists to figure out how to utilise these tools best in their work \cite{doerigNeuroconnectionistResearchProgramme2022}. The main requirement for these models to work has been collecting enough high-quality data for training. Especially, scaling into very large models has required scaling the sample size as well. A few large EEG datasets exist in different domains, for example diagnosis \cite{dzianokPEARLNeuroDatabaseEEG2024} and BCI \cite{jayaramMOABBTrustworthyAlgorithm2018}. None of these holds as much potential for insight into cognition as the THINGS initiative \cite{hebartTHINGSDatabase8542019}. THINGS contain 12 or more naturalistic images of 1,854 different objects with plenty of semantic details described for each object category including wordnet relationships and human-rated high-level categories. Crucially, recordings of humans watching these images now exist both for fMRI, MEG and EEG \cite{giffordLargeRichEEG2022b}. The EEG dataset by \cite{grootswagersHumanEEGRecordings2022} has 46 subjects all viewing 22,248 images, resulting in 1,023,408 total trials, which we believe warrants an interest from the computational cognitive neuroscience community. Despite this, not much work has been done in trying to utilise the large dataset for bigger DL models. This is why we, in this work, aim to take the first steps in exploring and benchmarking a simple task derived from this dataset using different popular DL models and comparing them to the best linear modelling techniques. 

\section{Related works}
The THINGS EEG dataset presents a unique challenge in decoding because of its experimental constraints. To practically let each subject watch 22,248 images, they are shown in rapid succession. Specifically 50 ms stimulus time and 50 ms off duty. That means ten images are shown per second. This presumably limits the amount of information processing the subject is doing per image, which is part of why this work needs to be benchmarked. An important constraint then is the short time window of the relevant signal. Crucially, this makes a frequency representation of the data less feasible. Band power representation has otherwise shown to be greatly suited for DL on EEG \cite{zhang2024torcheeg}. Therefore, we are tasked with analysing the untransformed time representation, which limits the possible comparisons for this work. The closest work in this area, to our best knowledge, comes from the BCI community, where benchmarking model performance is important. The MOABB project \cite{jayaramMOABBTrustworthyAlgorithm2018} lays out a variety of BCI datasets of various sizes and collects them for standardised testing. Interestingly, according to their leaderboards, almost all BCI applications are best solved using a linear decoder pipeline. It is worth noting that BCI tasks are often chosen specifically because they are easily solved by a linear model. For example, a binary motor imagery task often consists of imagining moving either the right or the left arm. This should create a clear lateralisation in the signal, which is easily picked up by a linear model. We, therefore, should include a linear modelling approach in our study to compare deep learning models. 

Recently, the THINGS MEG dataset was used for real-time image reconstruction \cite{benchetritBrainDecodingRealtime2024}. While MEG and EEG measure very similar signals, the MEG THINGS dataset did not have the ten images per second setup, making comparisons less easy. They used a convolutional model with residual connections introduced in \cite{defossezDecodingSpeechPerception2023} to get a 7X improvement to a linear baseline model, although their task included the original 1854 object categories, which is different to the task used here. 

\section{Methods}

\subsection{Task}
To keep in line with more traditional decoding paradigms, we introduce a novel binary classification task created from a subset of the EEG THINGS dataset. We want the task to be cognitively relevant and non-trivial to solve while still being possible. We chose the task of separating living vs nonliving things - specifically objects that are easily manipulated by hands. It is an established finding that this high-level categorisation can be separated by visual stimulation using fMRI \cite{mahonCategorySpecificOrganizationHuman2009} and EEG \cite{ahmadi-pajouhFractalbasedClassificationHuman2018}. We chose 429 object concepts (of the original 1854) that fit into these two categories from the human-labeled categories provided by the original THINGS paper \cite{hebartTHINGSDatabase8542019}. Specifically, we used the "Top-down Category (manual selection)". The categories chosen for handleable objects are: "tool", "sports equipment", "musical instrument", "electronic device", and "weapon", and for living the categories are: "animal", "animal, bird", "animal, insect", "animal, food", "body part". Note that "animal, food" refers to animals that could be eaten (e.g. "eel") but the picture is of a living animal, not food. It is worth noting that while the overall classes are completely balanced, the object categories used to construct them are not. This is not a purposeful construction, but a byproduct of the human-labeled categories of naturalistic images. An analysis of object-level performance is provided later. 
With 12 repetitions of each label and 46 subjects, that gives the task 236,808 total number of trials. We judge this to be adequate to start testing DL models with reasonably large parameter sets. We also created a single-subject task, in which each subject"s data was treated separately, resulting in 5,148 samples per subject.

\subsection{Preprocessing}
We kept preprocessing minimal \cite{delormeEEGBetterLeft2023}. The original data was referenced inconsistently in some subjects, we fixed this by re-referencing the data to the Cz electrode and included all other 63 electrodes. We applied a 1-40 Hz bandpass filter and downsampled the data to a 100 Hz sampling frequency. We then applied a baseline from 200 ms before stimulus onset and up to stimulus onset. Then only the 500 ms post stimulus onset was considered, resulting in a sample shape of 63 electrodes and 50 time points. The data was z-score normalised at each channel for each trial. 

\subsection{Training}
All models were trained using the same protocol, without any specialised hyperparameter tuning. The batch size was kept at 128 the optimiser was a standard stochastic gradient descent with a momentum of 0.9. The learning rate was on a cyclic schedule \cite{loshchilovSGDRStochasticGradient2017}, which makes it gradually decrease with "warm restarts". We kept the cycle parameters of the original paper, making starting cycle length $T_0=15$, and the cycle multiplier $T_{mult}=2$. This learning rate scheduling lets us avoid tuning the learning rate for each model. We trained the models for 945 epochs, giving us six warm restarts during training. For the single-subject task, we only trained models for 460 epochs, giving five restarts. In the original EEG THINGS paper, they ended their recording session with an inbuilt test split, showing a repeat of one of the twelve images in each category. We chose not to use this, as there could be potential cofounders with only using the last trials as the test set. Instead, we split the set randomly and tested on 20\% of the original data. All models are trained on one NVIDIA V100 graphics processing unit with 32 GB of memory, which was enough to contain the full dataset and models. 

\subsection{Models}
Six different model architectures were tested for this experiment, representing widely different modelling approaches for sequences: convolution, graph, recurrent and transformer. Representing convolutional models, we chose the popular EEGNet \cite{lawhernEEGNetCompactConvolutional2018}. A compact and efficient way to handle time series, convolutional kernels are applied across the signal and summarised by max-pooling layers. Then representing graph neural networks, we chose the dynamic graph convolutional neural network (DGCNN) \cite{songEEGEmotionRecognition2020}. Graph neural networks encode the electrode montage as a graph structure, modelling the electrode relationships as a non-euclidian space. Since EEG electrodes do not inherently have a graph structure, they let the adjacency matrix be a learnable part of the network, hence the name "dynamic". The DGCNN uses spectral graph convolution to transmit information across the graph. Then, for a recurrent network, we chose the long-short term memory (LSTM) architecture for EEG described in the book \cite{zhangDeepLearningEEGBased2021}. The LSTM works by establishing consecutive recurrent layers with gating for long and short-term relationships across the time series. Recently, the transformer model has shown its prowess in NLP, but it is not clear how it works on more standard time-series data structures. We decided to test the standard transformer model as described in \cite{vaswaniAttentionAllYou2017} using the attention mechanism to find relationships across the time series. Then, because the vanilla transformer is not standard for time series analysis, we also included the EEG Conformer \cite{songEEGConformerConvolutional2023}, which combines a convolutional block to encode the time series to make it better suited for the transformer model. The implementations for all these models were based on the TorchEEGEMO \cite{zhang2024torcheeg} and Braindecode PyTorch module \cite{HBM:HBM23730}, to ensure these were as standardised as possible. 

Only small changes were made in how dropout was handled. Three versions of each model were created to allow for diversity in model size. One "small", one "medium" and one "large" for each model type chosen. With the size of the training set as a guide ($\sim200,000$ samples), we aimed for the small model to be underparameterised ($\ll$100,000), the medium-sized model to have a similar size ($\sim100,000$) and the large model to be overparameterised ($>100,000$). Some models differed from this, to avoid straying too far away from the original design. The number of trainable parameters for each model can be seen in Table 1 and details of the implementation can be seen in the provided code. The dropout rate was changed for each model size, so that small models had a dropout of $0.25$, medium-sized models had $0.5$ and large models had $0.75$, to prevent overfitting. Additionally, weight decay regularisation was applied equally to all models, with $\lambda=0.001$. For the single-subject task, only small models were used and a dropout rate of $0.5$ was used.

For a linear baseline model, we chose the Common Spatial Patterns + Linear Discriminant Analysis (CSP + LDA) \cite{kolesQuantitativeExtractionTopographic1991}. This method is an adapted LDA model specifically for EEG event-related (ERP) signals. CSP + LDA works by identifying separatable spatial patterns as orthogonal waveforms in the EEG signal and treating them as features for an LDA. This is the method that performed best in most of the MOABB \cite{jayaramMOABBTrustworthyAlgorithm2018} leaderboards of BCI tasks, and typically works very well in clearly separatable conditions, see for example \cite{alotaibyEpilepticSeizurePrediction2017}. All models were first tested on a trivial task on the THINGS dataset with the same preprocessing, data structure and training method and achieved $>95\%$ accuracy. The trivial task was a binary classification task of identifying the subject of the trial. This shows that all models can learn by using this methodology. 

\setlength{\tabcolsep}{6pt}
\begin{table}[h]
    \caption{Number of trainable parameters in each model type and size}
    \centering
    \begin{tabular}{l|rrr}
        \toprule
        Model type & Small & Medium & Large \\
        \midrule
        EEGNet & 1,888 & 11,504 & 132,096 \\ \midrule
        LSTM & 3,902 & 98,402 & 1,161,002 \\ \midrule
        DGCNN & 12,527 & 107,563 & 1,049,763 \\ \midrule
        Transformer & 3,090 & 141,866 & 1,144,834 \\ \midrule
        Conformer & 36,026 & 164,906 & 1,404,946 \\
        \bottomrule
    \end{tabular}
    
    \label{tab:model size}
\end{table}

\section{Results}

\subsection{Cross-subject task}
The accuracy, precision and recall of each model on the cross-subject task test set are reported in Table 2 alongside their training time. Since the learning rate scheduling cyclicly resets performance, the performance was measured as the top recorded accuracy between the last 5 epochs before a warm reset. We remind the reader, that the task is a binary classification and the chance level is at 0.5. Best results are highlighted with \textbf{bold} text. 

\setlength{\tabcolsep}{3pt}
\begin{table}[h]
    \centering
    \label{tab:group results}
    \caption{Cross-subject performance metrics and training time.}
    \begin{tabular}{l l r r r r}
        \toprule
        Model & Size & Accuracy & Precision & Recall & Training time \\
        \midrule
        EEGNET & small & \textbf{0.5710} & 0.5713 & 0.5717 & \textbf{2h 47m 39s} \\
         & medium & \textbf{0.5710} & 0.5713 & 0.5711 & 3h 37m 56s \\
         & large & \textbf{0.5710} & 0.5701 & 0.5702 & 7h 57m 33s \\ \midrule
        LSTM & small & 0.5665 & 0.5677 & 0.5703 & 2h 54m 59s \\ 
         & medium & 0.5641 & 0.5711 & 0.5713 & 2h 54m 27s \\
         & large & 0.5665 & 0.5697 & 0.5705 & 4h 7m 41s \\ \midrule
        DGCNN & small & \textbf{0.5708} & \textbf{0.5715} & 0.5729 & 2h 53m 39s \\
         & medium & 0.5681 & \textbf{0.5715} & 0.5736 & 2h 54m 8s \\
         & large & 0.5660 & 0.5657 & \textbf{0.5788} & 2h 55m 0s \\ \midrule
        Transformer & small & 0.5041 & 0.4938 & 0.5021 & 14h 3m 40s \\ 
         & medium & 0.5047 & 0.4991 & 0.5043 & 14h 8m 24s \\
         & large & 0.5042 & 0.4997 & 0.4994 & 14h 8m 24s \\ \midrule
        Conformer & small & \textbf{0.5710} & 0.5713 & 0.5718 & 6h 25m 28s \\ 
         & medium & \textbf{0.5710} & 0.5713 & 0.5711 & 13h 12m 54s \\
         & large & \textbf{0.5710} & 0.5710 & 0.5710 & 15h 21m 1s \\ \midrule
         CSP + LDA & linear & 0.4970 & 0.5067 & 0.49320 & 0h 31m 21s \\
        \bottomrule
    \end{tabular}
\end{table}

We see that multiple models converged to a top performance measure of 57.10\% accuracy. Both EEGNET, the Conformer and the DGCNN reached this mark or within 0.01\% of it. Similarly, we see precision and recall around 57.15\% in those same models, showing how there was no clear bias to either false positives or negatives. A notable exception is the DGCNN, which reached 57.88\% in recall, which is marginally higher. The LSTM reached a peak accuracy of 56.65\%, making it almost as good. The vanilla transformer was by far the worst performing DL model, not able to reach above chance level. Crucially, the CSP + LDA linear model did also not reach more than the chance level. For further analysis, we identify the best-performing models as the EEGNet, DGCNN and the Conformer. 

\subsubsection{Training speed}
The small versions of EEGNET, LSTM and DGCNN all trained in under 3 hours. The medium-sized LSTM and the medium and large versions of DGCNN also cleared this mark, showing how scaling the graph neural network did not slow performance. EEGNET did slow down with bigger parameter sets, and the LSTM also slowed down with its large version. All these models were somewhat shallow, with only a few layers, which might explain how they were able to train this fast. It might also show, that reaching a breaking point in too many parameters per layer slows down training immensely. The transformer model was the slowest to train ($>$14h), with no improvement with smaller parameter sets. This is likely due to its need to find relationships across the time series, resulting in poor scaling in the sequence dimension. The Conformer was in between these at 6h 25m, showing how introducing convolutional kernels to the transformer model helps both with efficacy and efficiency.

\subsection{Single-subject task}

\setlength{\tabcolsep}{3pt}
\begin{table}[h]
    \centering
    \label{tab:ss results}
    \caption{Average best performance on the single-subject task.}
    \begin{tabular}{l|rrr}
        \toprule
        Model Type & Accuracy & Precision & Recall \\
        \midrule
        EEGNet & \textbf{0.6098} & \textbf{0.6107} & \textbf{0.6117} \\ \midrule
        LSTM & 0.5669 & 0.5677 & 0.5721 \\ \midrule
        DGCNN & 0.5663 & 0.5670 & 0.5680 \\ \midrule
        Transformer & 0.5010 & 0.4988 & 0.5034 \\ \midrule
        Conformer & 0.6037 & 0.6042 & 0.6097 \\ \midrule
        CSP + LDA & 0.4983 & 0.4983 & 0.5009 \\ \midrule
    \bottomrule
\end{tabular}
\end{table}

Table 3 reports the average best performance across subjects on the single-subject task. We extract this by finding the best test performance during the average of the last five epochs per restart cycle. We see that the best-performing model is the EEGNet reaching 61\% in accuracy, precision and recall. The Conformer model is close in performance, hovering around 60.5\% in all three measures. Neither the transformer nor the linear CSP + LDA was able to learn the dissociation, getting around the chance level in all measures. The LSTM and the DGCNN achieved similar or worse performance, compared to the cross-subject task, suggesting that the smaller dataset made it harder to generalise to the test set effectively. 
\begin{figure}[h]
    \includegraphics[width=\textwidth]{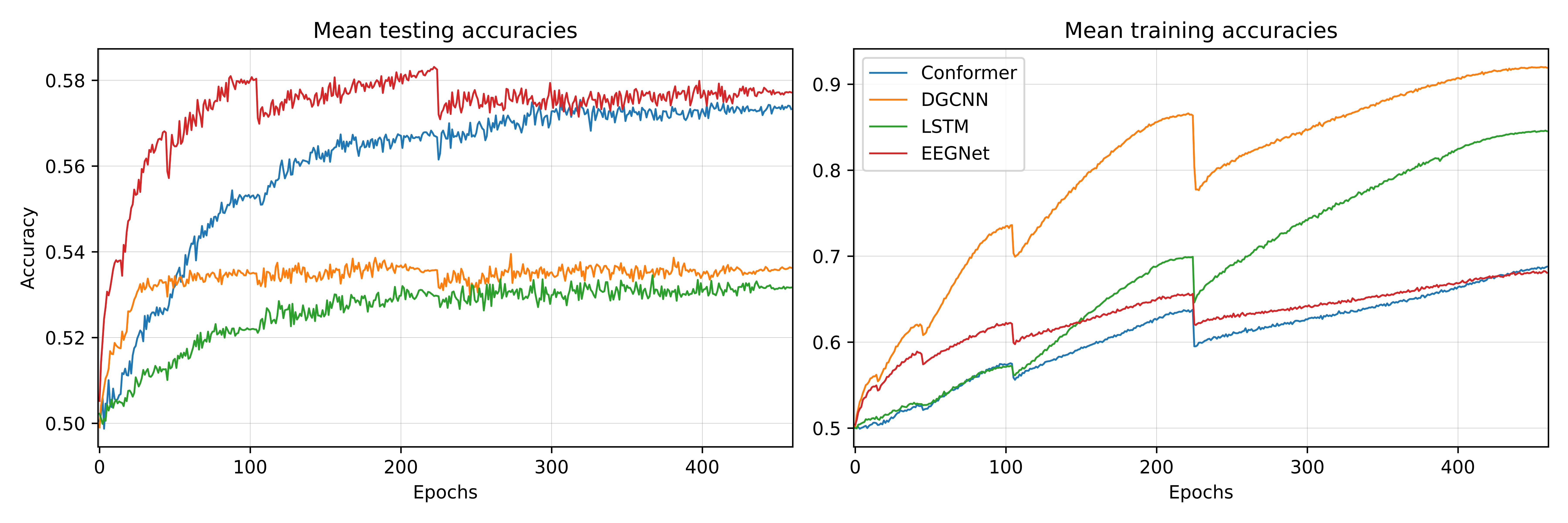}
    \caption{Mean accuracies during training} 
    \label{training accs}
\end{figure}
Fig. 1. shows the average performance during training. Note that the numbers in the table and the figure are not directly compatible as different subjects might reach the best performance during different training cycles. Here we see the propensity of all models to overfit on the smaller dataset. We can also see that EEGNet reaches peak performance much faster (100 epochs) whereas the Conformer needs 300 epochs of training. The LSTM and DGCNN were more susceptible to overfitting getting lower testing performance but higher training accuracy.

\subsection{Comparative analysis on stimulus labels}
\begin{figure}[h]
    \includegraphics[width=\textwidth]{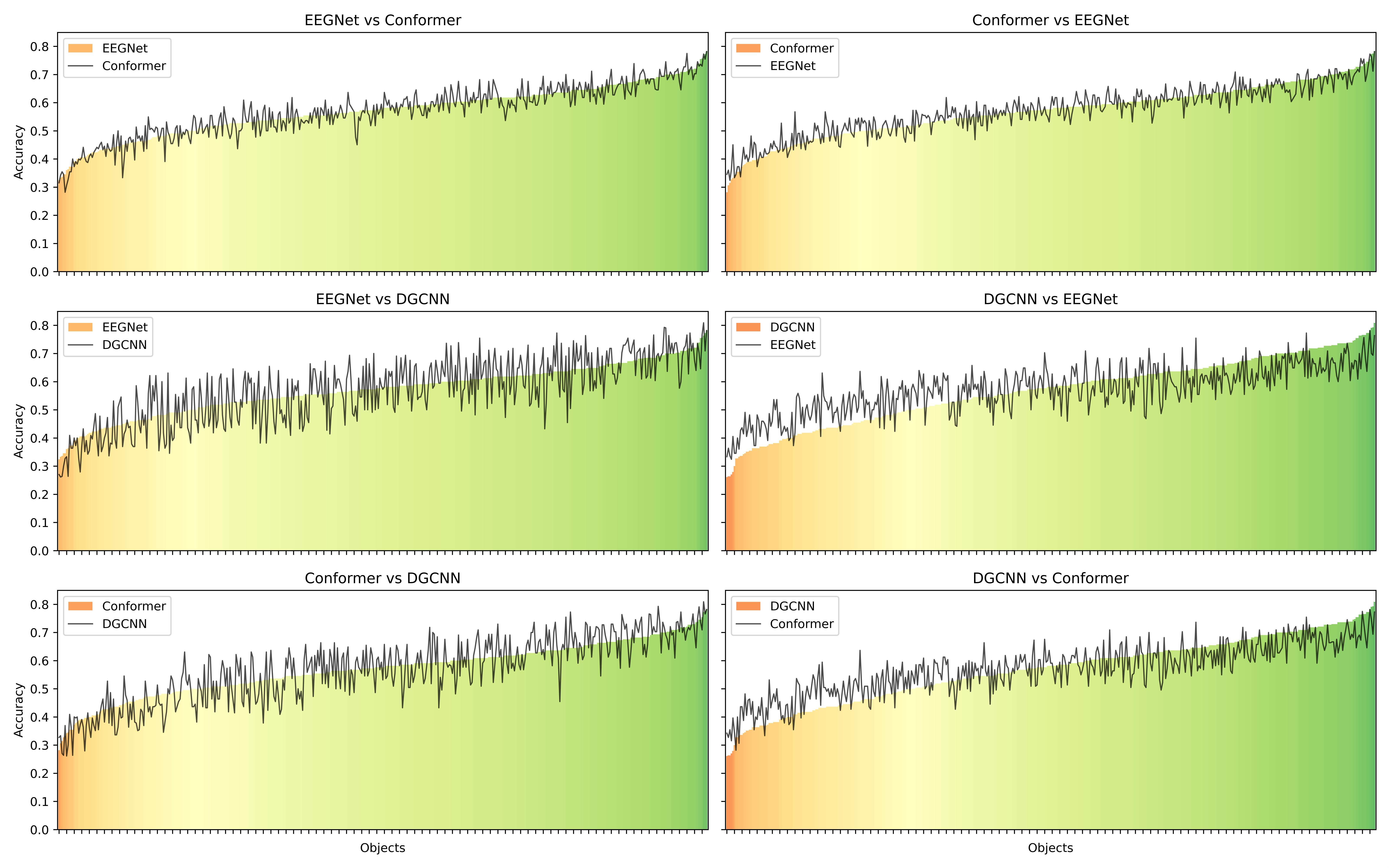}
    \caption{EEGNet, Conformer and DGCNN performance compared on different object labels. Gradient barplot (background) is ordered according to average test performance on each object, with one slice representing a single object category. The colour gradient maps accuracy, with red being 0 and green being 1. The overlaid line represents the compared model and follows the same order, to reveal potential differences.}
    \label{accs_comparison}
\end{figure}

Motivated by the above results of a somewhat stabilised performance across different model architectures, we ran a more detailed analysis of the individual object labels and the top-down labels, both provided in the original THINGS initiative work. We start with a comparative analysis of all object categories, see Fig. 2 for visualisation. Since multiple models achieved similar performances, it is worth investigating if they are classifying the same image categories with similar success. Here we use the cross-subject models and stick to the small version of each model, that achieved 57.1\% accuracy.

\subsubsection{Object labels}
Looking at Fig. 2. we see that all models follow roughly the same trend of which object categories that were labelled correctly. Especially the Conformer and EEGNet show very similar behaviour on the test set, whereas DGCNN shows the same overall distribution but with higher variance for individual object labels. In order to verify that these indeed have similar performance, we run a paired t-test (428 degrees of freedom) on the accuracies for each object. None of these showed a significant difference: Conformer vs EEGNet - $t = 1.116$ $ p = 0.265$ , EEGNet vs DGCNN - $t= 1.175$ $p= 0.241$, Conformer vs DGCNN - $t= 0.794$ $p= 0.428$. We also note that there is some difference among object labels. Best-performing objects have around 70\% accuracy, while the worst-performing labels sit under 30\%. We note that there seem to be a few objects that give a particularly bad performance, as seen in the small dent on the far left side of each subplot in Fig 2. 

\subsubsection{Top-down categories}

\setlength{\tabcolsep}{5pt}
\begin{table}[h]
    \centering
    \label{tab:Objects}
    \caption{Best models average accuracy on top-down categories}
    \begin{tabular}{l |l| r |r}
        \toprule
        Label & Top-down category & \#Objects & Accuracy \\
        \midrule
        Alive & animal & 113 & 0.5511 \\
         & body part & 34 & 0.4669 \\
         & animal, bird & 25 & 0.5230 \\
         & animal, food & 20 & 0.5006 \\
         & animal, insect & 17 & 0.4196 \\ 
         & people & 5 & 0.5607 \\ \midrule
         & all & 214 & 0.5185 \\ \midrule
         Non-living & tool & 59 & 0.6270 \\
         & sports equipment & 51 & 0.6097 \\
         & electronic device & 43 & 0.6432 \\
         & musical instrument & 33 & 0.6143 \\
         & weapon & 29 & 0.5881 \\ \midrule
         & all & 215 & 0.6176 \\ \midrule
         Total & & 429 & 0.5710 \\
        \bottomrule
    \end{tabular}
\end{table}

\begin{figure}[h]
    \includegraphics[width=\columnwidth]{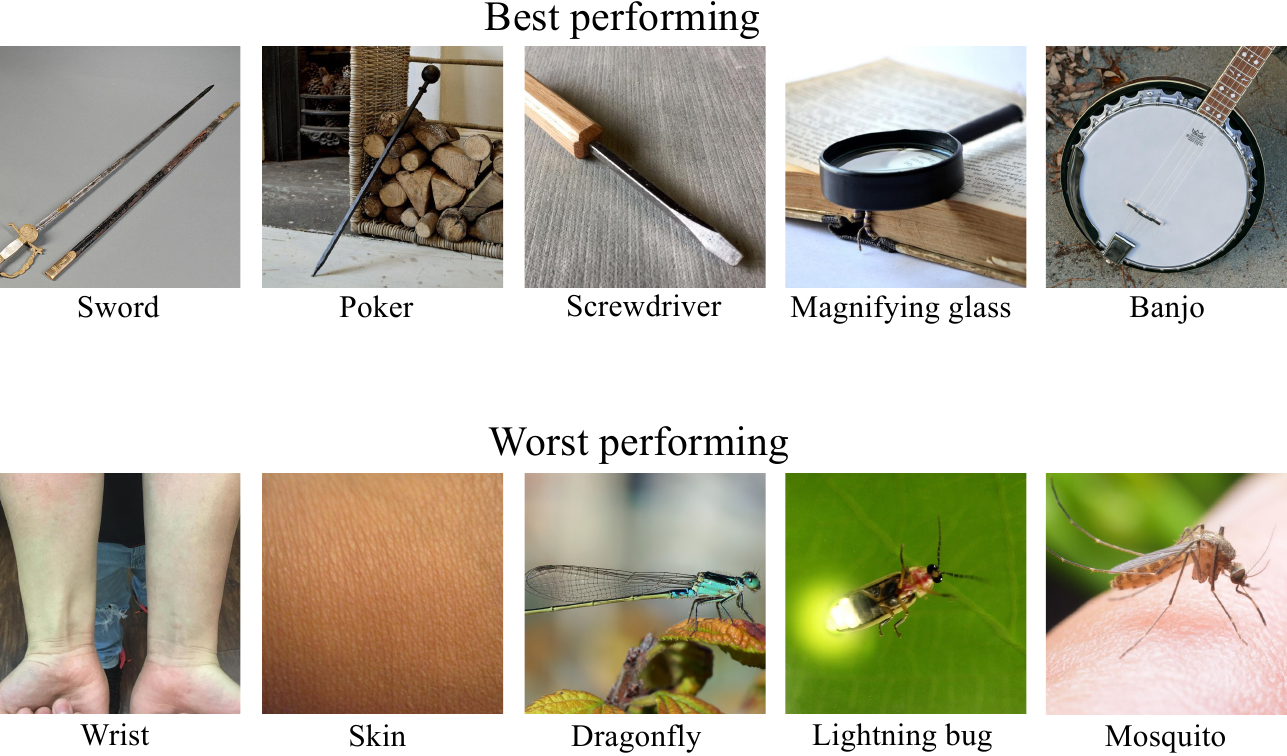}
    \caption{Example images of the best and worst performing object categories, taken from the THINGS initiative available at: https://osf.io/jum2f/.} 
    \label{training accs}
\end{figure}

Lastly, we look at the performance divided into the top-down categories, that we originally chose to make the animacy labels. It is important to note that these categories are not balanced, meaning some categories like "animal" hold over 100 different objects, whereas "people" only contain five. In Table 4, we report the average accuracy of the small version of EEGNet, DGCNN and the Conformer on the test set, stratified by these top-down categories. First, looking at the "alive" categories, it is clear that especially the two categories "body part" (46.7\%) and "animal, insect" (42\%) are underperforming compared to the others. While one could argue they are underrepresented in the data, as both have a low number of objects (34 and 17), we also see that the "people" category has the highest accuracy (56\%) among the "Alive" categories, while being the worst represented (5). Another possible explanation is these images have different image compositions or low-level image features (e.g. contrast, colours) to the other categories, resulting in highly different EEG signals, see Fig 3. for example images. This explanation would require a more detailed analysis of the original images. The last explanation is the cognitive one, where insects and individual body parts are simply not reliably treated as "living" in the same way that other animals and people are. Again more analysis would be needed to confirm or deny this hypothesis in later work. 

We then look at the "Non-living" categories and we notice two main things. Firstly, the distribution of accuracies is much more even, all of them hovering around 60\%, maybe with the exception of "electronic devices" sitting at 64.3\%. This even spread could simply be from the much more balanced number of objects in each category, reducing the risk of model bias towards specific object categories. It could also reflect a higher degree of similarity in the neural representation in these categories than seen for living objects. The second thing to notice is that the average accuracy among "non-living" is almost 10\% higher than that for "alive" categories. We also see this reflected in Fig 3., where the top five performing categories are "non-living" and all worst performing five categories are all labelled "alive". Possibly this shows that there was higher homogeneity across the objects chosen as "non-living" since these were specifically picked to be "handleable" objects, that is objects one could interact with using one's hands for some purpose. Indeed, it has long been known that such affordances drive reliable motor cortex activation \cite{maranesiCorticalProcessingObject2014}, which could be driving the higher performance in these objects compared to animals. We provide possible next steps in the Future Work section below. 

\section{Discussion}
In this work, we tested 5 different DL architectures and a popular linear model on a binary classification task of separating living vs. non-living entities, which was created using the EEG THINGS dataset. Interestingly, the linear model did not manage to find the dissociation between these two conditions. Instead, every DL model, except the vanilla transformer, managed to achieve $>$56.5\% accuracy on both the cross-subject and the single-subject task, showing a reliable decoding result. Firstly, this seems to demonstrate that in some decoding tasks applying a nonlinear model is necessary to find patterns between stimulus and brain activity. Secondly, we show here the feasibility of decoding high-level object features from rapidly shown images (10 hz), even though the signal might be overlapping between trials. We did not find that the specific model choice mattered greatly. Vastly different architectures like the Conformer, recurrent and graph neural networks all achieved the same peak performance. Perhaps suggesting a cap in possible performance on this task around 57.10\% accuracy. We leave it as an open challenge for future work to achieve better results using the same setup. 

We also ran an analysis on the similarity of performance on different object labels in the three best-performing models, EEGNet, DGCNN and Conformer. We found that the differences between models were not significant, but also not identical. Especially the convolutional neural network EEGNet and the convolution+transformer architecture in the Conformer have similar performance. The EEGNet and the Conformer also performed best in single-subject tasks, where the sample size was much lower. One interesting observation is then that the convolutional filters seem to be an effective representation of EEG signals when the number of samples is on the lower side. Even with the larger parameter (see Table 1) set of the Conformer (36,000) compared to the DGCNN (12,500) and LSTM (3,900), overfitting did not disrupt performance. Possibly, the convolutional filters act as an information bottleneck, squeezing the features through a smaller parameter set and decreasing the risk of overfitting. 

We did only test this on a subset of the original full dataset, and on a single binary classification task. However, this result is shown robustly across different domains as an achievable task and the fact that the linear model was not able to solve it in this case seems to suggest the necessity to use deep models in some circumstances. As discussed above the THINGS dataset presents a unique challenge, as it is uncertain how much information is present in the EEG signal, when images are presented in such a rapid fashion. It is certain that having overlapping stimuli in the signal will decrease the signal-to-noise ratio for single trials. This is compensated for, by having many more trials. Our results suggest that DL models are well positioned to perform decoding tasks in this setting of many trials of lower signal quality. 

\subsection{Future Work and Conclusions}

Much future work is needed in this area. Here we chose to test different model types and sizes as the main independent variables. However, other choices would be equally interesting to test. We suggest looking at different preprocessing procedures like filtering, artefact removal and normalisation. Not much research has been done on decoding the performance of deep learning models on EEG or MEG signals, which leads to relying mostly on conventions. We would also suggest trying to decode other high-level or low-level object features. The THINGS initiative provides many different possible image labels, which could be used as targets for decoding tasks.

Another important avenue for future research is to develop a better understanding of model behaviour. We ran a simple comparative study here, but much more work is needed to understand how DL models represent and analyse EEG signals. Our analysis of object labels showed that models had quite different performances on different top-down categories of objects. We saw how convolutional models were successful in our study. One could use visualisation techniques to understand what parts of the signal are driving the performance of the models. Similarly, the DCGNN are learning a graph representation of the EEG electrode relationships during training. We suggest that looking at this graph might give insight into how the model has learned to represent the signal and hint at a functional connectome. 

While this work replicated a well-known finding in the cognitive neuroscience literature, that the animacy of seen objects can be decoded from brain signals, this is an important step for incorporating DL techniques into the toolkit of cognitive neuroscience. Artificial neural networks hold much potential but need to be explored more systematically in new and older settings, to test where they are useful compared to traditional methods. Replicating known findings using new methods is a reliable and necessary preliminary step to test the method's overall utility. With the entrance of bigger and more complex datasets like the THINGS initiative, using DL models to perform neural decoding in settings where linear models simply cannot capture the relationships is a likely next step for using functional neuroimaging to advance our knowledge of brain representations.

\bibliographystyle{splncs04}
\bibliography{references}
\end{document}